# Simultaneous localization and mapping by using Low-Cost Ultrasonic Sensor for Underwater crawler

Trish Velan Dcruz
School of Electronics Engineering
*Vellore Institute of Technology*
Chennai, India
trish.dcruz2019@vitstudent.ac.in

Cicero Estibeiro
School of Electrical Engineering
*Vellore Institute of Technology*
Chennai, India
ciceroraimundo.estibeiro2019@vitstudent.ac.in

Anil Shankar
School of Electronics Engineering
*Vellore Institute of Technology*
Chennai, India
anilshankar.s2019@vitstudent.ac.in

Mangal Das
School of Electronics Engineering
*Vellore Institute of Technology*
Chennai, India
mangal.das@vit.ac.in

*Abstract - Autonomous robots can help people explore parts of the ocean that would be hard or impossible to get to otherwise. The increase in the availability of low-cost components has made it possible to innovate, design, and implement new and innovative ideas for underwater robotics. Cost-effective and open solutions that are available today can be used to replace expensive robot systems. The prototype of an autonomous robot system that functions in brackish waterways in settings such as fish hatcheries is presented in this research. The system has low-cost ultrasonic sensors that use a SLAM algorithm to map and move through the environment. When compared to previous studies that used Lidar sensors, this system's configuration was chosen to keep costs down. A comparison is shown between ultrasonic and lidar sensors, showing their respective pros and cons.*

*Index Terms – SLAM, AUV, Ultrasonic sensor, Lidar.*

## I. INTRODUCTION

The aim of this project is to provide a useful module for making underwater maps and using a crawler to navigate the environment. It is a challenge to reliably localise and navigate in extremely unstructured underwater settings. To guarantee that accurate and repeatable measurements are being recorded for surveying and other uses, it is crucial to know the location and distance the autonomous underwater vehicle (AUV) has travelled. Simultaneous localization and mapping (SLAM) techniques and sensor fusion are two solutions implemented to achieve this goal. Yet, as newer techniques have mostly been used in land and air environments, there is still opportunity for utilising these strategies in underwater circumstances. In the past two decades, remotely operated vehicles (ROVs) and autonomous underwater vehicles (AUVs) have seen tremendous advancement. The operator of a ROV controls it from a station on the surface. Because of the tethered connection, they are able to control sophisticated instruments and offer high bandwidth for communication. The majority of visual inspections of dams, pipes, tunnels, and structures are conducted using smaller ROVs [1]. They are also used in fisheries, water quality assessments, fish habitat studies, and fishing. The completion of set tasks can be branched out with the use of various sensors to develop several applications. Underwater probes for water quality (which detect elements including temperature, salinity, depth, turbidity, and chemical composition) [2] are one type of sensor used to monitor the environment.

Moreover, acoustic sensors are used for measuring underwater noise, detecting objects, estimating biomass, and imaging in both a crude and precise manner. GPS is ineffective in submerged situations. Instead, networks of transducers and receivers are employed to triangulate location using acoustics for communication and localization [3]. In this paper, we present solutions to two main challenges in underwater robots. First, we focus on the localization of the underwater crawler and then the application of navigation in a real-time environment in the Robotic Operating System (ROS). SLAM algorithms are used to achieve autonomous navigation. These algorithms aid in the determining the robot's position as it moves through unexplored areas. The quality of the generated map is enhanced as the robot continues to explore the unknown area. We also provide a comparison using SLAM between LIDAR and cost-effective ultrasonic sensors, respectively.

## II. Simulation Software

### A. ROS

ROS (Robot Operating System) is a software application used for developing and simulating robots. It provides a standard communication protocol for exchanging messages between different parts of a robotic system. ROS is widely used in both academia and industry and also provides a range of tools for visualization, simulation, and testing, which makes it easier for developers to develop, debug, and test their robotic software.

B. Gazebo

Gazebo is a popular 3D simulation environment for robotics that is integrated with ROS. It provides realistic physics and supports various robot models. ROS offers tools and libraries for building complex robotic systems and enables users to control robots in simulation and receive sensor data. ROS provides a way to interface with Gazebo, allowing users to control robots in simulation and receive sensor data as if the robot were operating in the real world. This allows researchers to evaluate algorithms in different scenarios and environments without requiring physical experimentation.

C. RViz

RViz is a 3-D visualisation tool that works with ROS. It lets users see and interact with 3D models and sensor data made by robots in real time. It can display the robot's kinematic model, and users can interact with it using various tools. Its flexibility, customizability, and ease of use make it a valuable tool for developing and testing robotics applications.

III. METHODOLOGY

A. SLAM

SLAM is a basic problem in robotics that has to do with finding the best way to get somewhere while avoiding obstacles. It takes into account the robot's limitations, like its size, speed, and turning radius, as well as the environment's features, like obstacles, terrain, and other things. SLAM requires sensors that can provide high-quality environmental data. The paper provides a comprehensive comparison of two commonly used SLAM sensors, ultrasonic and LiDAR, based on their performance and cost. The comparison of their performance is based on accuracy, environment reliability, range, and cost. The findings indicate that LiDAR is a superior choice since it offers a greater range with lower running time [4]. However, in our paper, the environment is enclosed and small, so we use ultrasonic sensors with a limited range to save cost. This paper describes the implementation of optimal path planning to determine the most efficient and accurate path. The path planning approach employed in this paper is based on Dijkstra's algorithm. It is important to note, however, that having access to the map is necessary in order to carry out path planning.

Global path planning aims to find the shortest route, while the local path planning is utilized to avoid objects and obstacles.

$$f(x) = a(x) + b(x)$$

Here $f(x)$ is the total valuation function. $a(x)$ is actual cost(value) from starting point to current point, $b(x)$ is estimated cost from current point to end point. $b(x)$ is variable and we have used Manhattan distance formulae in our algorithm.

$$D_{man} = |p_1 - p_2| + |q_1 - q_2|$$

here $(p_1, q_1)$ and $(p_2, q_2)$ are the coordinates to measure distance between two points.

$$\Delta X = X + \nu \Delta t \cos(\theta_t)$$
$$\Delta y = y + \nu \Delta t \sin(\theta_t)$$

the above two equations denote the respective increment of costs along the x and y axis. This equation was further calculated to obtain the robots trajectory for local path planning equation.

$$\Delta X_{t+1} = X_t + \nu \Delta t \cos(\theta_t)$$
$$\Delta y_{t+1} = y_t + \nu \Delta t \sin(\theta_t)$$

B. SLAM SENSORS

During the initial development of SLAM systems, range sensors such as acoustic and LiDAR were being utilized. Acoustic sensors were predominantly used for short-range and underwater applications. In 2005, Nguyen et al. introduced LiDAR-based SLAM which is capable of ranging at much larger distances with accuracy. However, its main limitations are the absence of feature extraction visual information [5].

3.1. Ultrasonic sensor

The ultrasonic sensor is a popular choice due to various advantages like lower power consumption (0.01-1 W), better underwater accuracy, simple design and low computational costs [6]. This type of sensor emits an acoustic wave at a specific frequency and detects objects by receiving their echo signals. In the case of ultrasonic sensors, the waves travel at the speed of sound through the air and bounce back after striking landmarks at the same speed. By calculating the time, it takes for the signal to travel from the object to the receiver, one can estimate the distance of the object [7]. Our robot is equipped with four ultrasonic sensors that have a range of 2m and can take 10 samples per second.

The observation model describes how the robot's sonars measure the environment. This model is used to recreate a map of the underwater environment.

3.1.1 Principle of Ultrasonic Ranging

The ultrasonic transducer emits ultrasonic waves and detects the echoes reflected from an object, either by using the same transducer in pulse-echo mode, or by using another transducer in pitch-catch mode [10]. In pulse-echo mode, the distance (d) between the ultrasonic transducer and the object can be determined by measuring the time (t) between the transmission of the signal and the reception of the echo signal, as shown in the equation.

$$D = ct/2$$

where $c$ is the sound velocity in the medium.

### 3.1.2 Ultrasonic Reflections

The sensor works on the principle of time-of-flight, which involves sending an echo signal and receives it after a specific time duration. However, the sensor's limitations become more apparent during real-world mapping due to the beam width causing uncertainty in sensing direction, a limited field of view and range, and errors resulting from reflections from walls. Using more sensors can also lead to crosstalk, where one sensor picks up the echo of another. Ultrasonic waves can be absorbed by soft materials, and the angle of incidence can affect the reflection of the waves. If the angle of incidence is less than 45 degrees, the waves may not be reflected back to the sensor, resulting in a lack of measurement data as seen in Fig. 1. This is a limitation of using ultrasonic sensors for object detection. Differentiating between environmental features, such as corners and planes, based on a single scan of the sensor is difficult. The speed of sound waves can also vary with temperature, but for this experiment, it is assumed that the robot operates in a controlled indoor environment with a constant temperature.

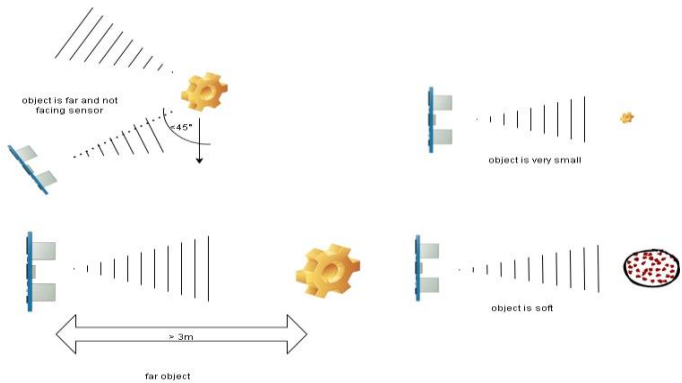

Fig.1. Reflections of the echo from the environment

### 3.2. Lidar Sensor

Detecting landmarks requires consideration of parameters such as the depth, scan angle, as well as accurately measuring the distance. In this context, LiDAR stands as its exceptionally accurate and has omnidirectional detection capability (360°), surpassing that of ultrasonic sensors, and allowing it to detect objects and landmarks from longer ranges. However, one major drawback is its high-power consumption, typically ranging from 50-200 W.

### 3.3 Drawbacks of Lidar in Underwater Applications

Compared to other remote sensing technologies, Lidar equipment suitable for underwater use can be quite expensive, making it unsuitable for certain applications. Additionally, Lidar is sensitive to water turbidity, which can negatively impact its accuracy by absorbing and scattering the signals. This issue can also lead to a reduction in the penetration of the Lidar underwater. Furthermore, surface reflection is another drawback of Lidar in underwater applications.

Scattering as well affects lidar's performance underwater. When the laser beam interacts with water molecules, suspended particles, and other objects, it scatters in different directions, making it difficult to estimate the precise location of the target [11]. When the laser beam hits a target, it reflects back, creating a signal that the lidar system detects. However, water has a high refractive index, causing the laser beam to bend as it passes through the water surface. This bending effect makes it difficult to determine the precise location of the target, leading to inaccurate depth measurements.

### 3.4 Comparison between SLAM sensors

| Properties | LIDAR | ULTRASONIC SENSOR |
|---|---|---|
| No of sensors | 1 | 4 |
| Samples per sec | 720 | 10 |
| Absorption underwater lidar (in m^-1) ultrasonic (in dB/m) | Blue-green light (400-500 nm): 8 to 16 Green light (500-550 nm): 4 to 8 Yellow light (550-600 nm): 2 to 4 Red light (600-700 nm): 1 to 2 | 1 kHz: 0.2 to 0.5 10 kHz: 2 to 3 100 kHz: 50 to 100 |
| Range under water | 1-10m | 0.01m – 2m |
| Accuracy(under water) | Better underwater accuracy for longer distances | Better underwater accuracy for short distances |
| Cost | Beyond ₹1,00,000 | 4 x ₹100 = ₹400 |

*3.4.1 Cost Effectiveness*

A critical factor in the selection of a SLAM sensor is its cost, particularly when aiming to build a low-cost autonomous robot. However, comparing sensors based on cost alone can be challenging as it depends on their accuracy and resolution, range and size [12]. Among all the SLAM sensors, ultrasonic sensors are relatively inexpensive, while LiDAR is the most expensive. However, the cost of LiDAR can vary significantly depending on its angular resolution, range resolution, range measurement and scanning angle [13].

*3.4.2 Measuring Range*

Using sonar sensors with a lower measuring range can be a practical choice in certain scenarios, such as in small and enclosed environments, where a shorter range is sufficient. It can also be a cost-effective solution. However, it is important to note that the range limitation of sonar sensors can limit the capabilities of the mobile robot, especially in larger and more complex environments. In such cases, LiDAR or other sensors with higher range capabilities may be necessary to ensure accurate range measurement and reliable mapping. For our experiment we have chosen to use sonars having a lower measuring range, as our environment has a small footprint as well as optimizing costs.

*3.4.3 Accuracy*

The accuracy provided by manufacturers may not be the same in all environments and conditions. Factors like noise, interference, and calibration can affect the accuracy of sensors. However, in this paper we consider the accuracies provided by the manufacturers such as HR304 (ultrasonic sensor) which has 3 mm, and LiDAR sensor systematic error is 50 mm, while the range of sensors is more than 10 m. Figure 3 shows the environment and map respectively utilized throughout the analysis. Regular house environment has been designed to showcase accuracy of obstacle detection of various types.

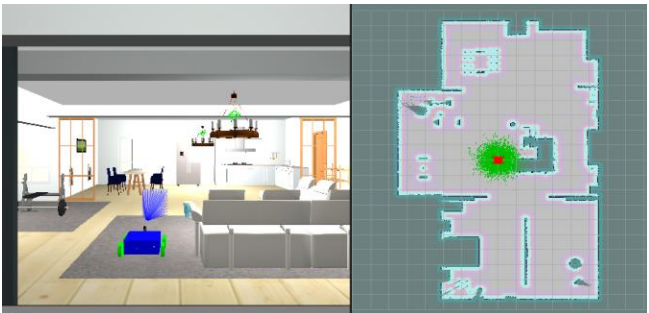

Fig. 3. Environment & Geometric Map

## IV. DESIGN

### A. Robot Model

The enclosed underwater robot has four wheels, a Lidar sensor, and four sonars, all equidistantly placed. Having four wheels provides the robot with the ability to manoeuvre and map the underwater area efficiently, especially in topographically complex environments. We can use either the Lidar or ultrasonic sensors to map the entire area and detect any potential obstacles. The four sonar sensors on the robot have a range of 0.01m to 2m, and ten samples are taken per second. The bot is controlled using Gazebo plugins (Steer drive controller, Teleop twist keyboard, Gmapping) built into the simulation software.

The robot was designed in a CAD simulation software as shown in Fig. 4. and is exported as URDF file, which are used to describe the robot model in the ROS framework. URDF files can be imported into simulation environments such as Gazebo to simulate the robot's behaviour in a virtual environment. Fig. 5. shows the prototype of the model after the URDF file was exported in Gazebo and allows us to check and test the basic functioning of the robot.

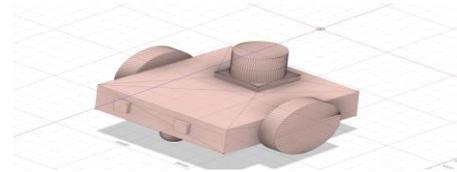

Fig. 4. Design

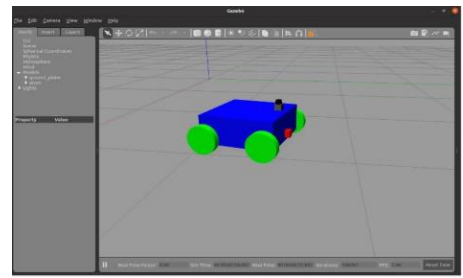

Fig. 5. ROS prototype

### B. Solid Works to URDF

SolidWorks is a software tool used for designing 3D models, allowing us to make precise measurements before implementing in ROS. Fig 6. shows the design process in SolidWorks, where the orientation and configuration of the sensors can be accurately represented in the simulation environment, allowing for more realistic and accurate testing of the robot's performance.

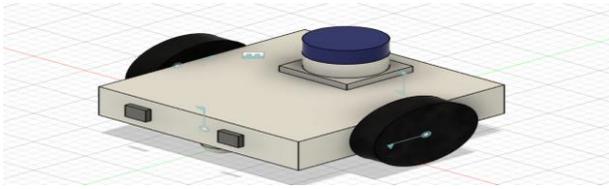

Fig. 6. Robot Model

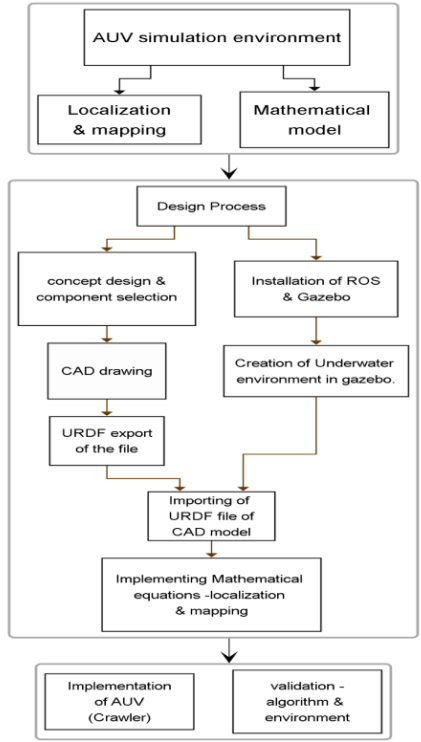

Fig. 7. Project Flowchart

## V. SLAM

### A. Methodology

SLAM involves constructing a map of an unknown environment while simultaneously localizing the robot within it. SLAM plays a crucial role in various robotic systems like autonomous vehicles, UAVs, and mobile robots. The problem with SLAM is that it requires the robot to estimate its position and orientation relative to the map, while simultaneously using sensor data to construct the map. This problem becomes more challenging in environments where there is significant uncertainty, like cluttered or dynamic environments. To address this challenge, SLAM algorithms utilize different sensors such as cameras, lidar, or sonar to collect data about the surroundings. The data is used to determine the robot's position and orientation and create a map of the environment. Different techniques such as probabilistic methods, graph-based methods, and machine learning are used in SLAM algorithms to estimate the robot's position and construct the map. SLAM has numerous applications in robotics, including exploration, surveillance, and mapping of unknown environments. SLAM is an essential technology for developing autonomous systems where accurate and reliable localization and mapping are vital for safe and efficient operation.

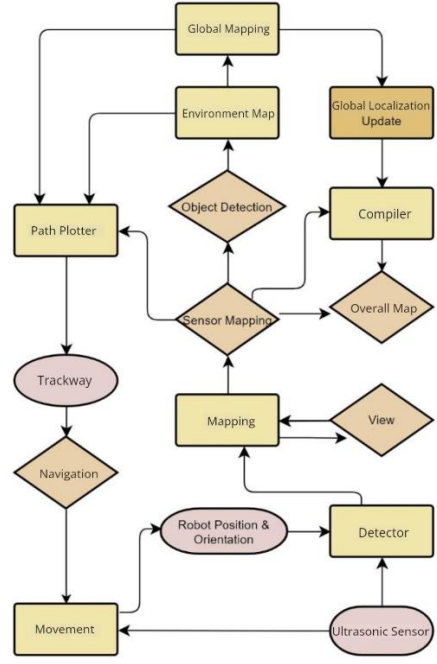

Fig. 8. Navigation Flowchart

### B. Localization

State vector: The state vector is a representation of the current estimation of the robot's orientation and position in the underwater environment. It comprises a vector of six elements, including the robot's position *(x, y, z)* and orientation *(roll, pitch, yaw)*, with the z-axis being constant.

Motion model: The motion model outlines how the robot's state evolves over time, taking into account factors such as currents, turbulence, and other environmental variables. However, the effect of these parameters is negligible in the crawler's working environment. Localisation of the crawler stabilizes when in motion as seen in figure 10 and is uncertain when stationary as shown in figure 9.

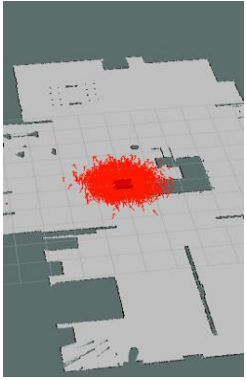
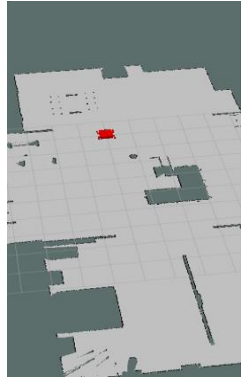

Fig. 9. Stationary      Fig. 10. In Motion

*C. Real-Time Navigation*

When comparing the LiDAR model with a single sonar model, the lidar model is superior in terms of range and providing a real-time map of the environment. The ultrasonic model can only map the environment within a range of 1-2 meters and detects changes in the surroundings only when it is within its range. The lidar model also offers a path planner, leveraging the entire surrounding map. However, for small-scale environments such as ponds or fish hatcheries, sonar models are more cost-effective, despite having a limited range. These environments do not require long-range detection as they are relatively small and do not extend beyond a few tens of meters. Figure 11 demonstrates the real-time mapping capabilities using a LiDAR sensor, while the left of figure 12 shows the range achieved by the ultrasonic sensor. As observed with LiDAR sensors, at a front sensing angle of 180 degrees, the LiDAR maps almost all of the front environment compared to the ultrasonic sensor in figure 12

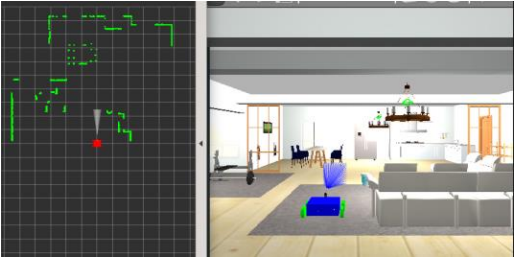

Fig. 11. Lidar

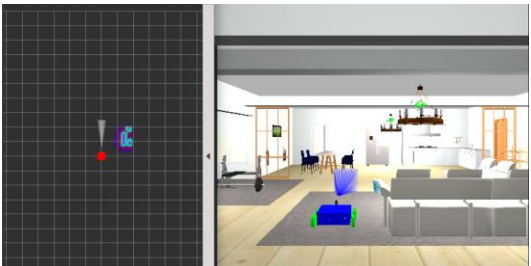

Fig. 12. Sonar

*A. Ultrasonic Mapping in empty environment*

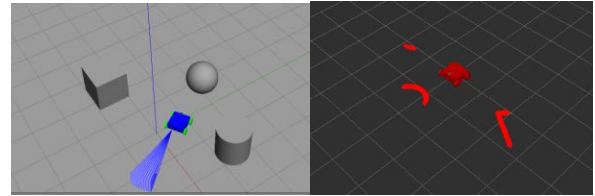

Fig. 13. Object Detection

Figure 13 describes how once scanned, the environment is mapped with current readings, and is only updated when the ultrasonic sensor encounters any change. Until the change is detected the last scan is saved on the current map. The observation model describes how the sonars measure the distance and reflectivity of the surrounding objects, which can then be used to construct a map of the underwater environment. within its range.

$$d = c * t / 2$$

Sound waves from sonar sensors spread out as they travel through the water, leading to a reduction in signal strength over distance. The beam spreading theta can be deduced by using the beam width w and the range r as follows:

$$theta = 2 * arctan(w / (2 * r))$$

The range of the ultrasonic sensor is based on the front facing direction of the crawler. Hence figure 14 gives the general range of the crawler based on the ultrasonic sensor's readings.

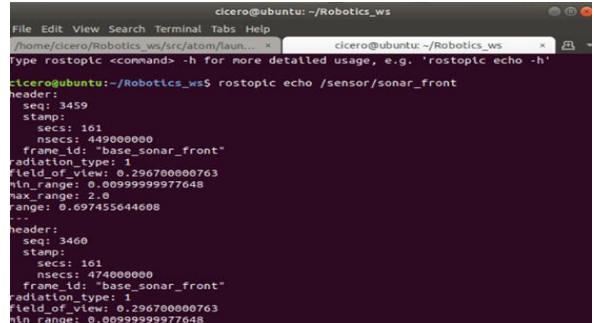

Fig. 14. Sonar Readings

## VI. COMPARATIVE ANALYSIS

The LiDAR sensor, although provides superior range as seen in all the images shown above, given the drawbacks and costs required, Lidar sensors aren't required for the purpose of navigation in our environment, given the practical working conditions. Having been able to accomplish the same task as a LiDAR sensor, Ultrasonic sensors are much more cost efficient and advantageous compared to laser scanning sensors when it comes to underwater applications as stated in previous parts of the paper.

## VII. CONCLUSION

This paper presents the successful implementation of an autonomous navigation system using ROS in a simulated environment created with Gazebo. Ultrasonic sensors and LiDAR are used to collect environment data, and a complete coverage path planning algorithm is used to ensure complete data collection. The navigation methods used in this paper include SLAM and path planning algorithms. The robot successfully generates a map of its environment and traverses through it, to reach the assigned destination without collision. In case of an obstacle, the path is rectified based on sensor feedback. The system enables the robot to navigate autonomously and collect terrain data in an unknown environment. Further software development and analysis could improve the precision of navigation, while full hardware implementation and calibration remain for future work.

### ACKNOWLEDGMENT

We would like to express our gratitude to those who have supported and contributed to the completion of this paper. First and foremost, we thank our guide Mangal Das for his guidance, encouragement, and valuable feedback throughout the research process. His expertise and support have been crucial in shaping this paper and improving its quality. Secondly, we thank VIT University for providing us with the facilities to enable us to complete our project. Additionally, we are grateful to the reviewers for their insightful comments and suggestions that helped us improve the quality of this paper. Without the support and contributions of these individuals, this paper would not have been possible.